\documentclass[10pt]{article}

\usepackage[letterpaper,margin=0.82in]{geometry}
\usepackage[T1]{fontenc}
\usepackage{lmodern}
\usepackage{microtype}
\usepackage{amsmath,amssymb}
\usepackage{booktabs}
\usepackage{tabularx}
\usepackage{array}
\usepackage{float}
\usepackage{enumitem}
\usepackage{graphicx}
\usepackage{xcolor}
\usepackage{xurl}
\usepackage{tikz}
\usetikzlibrary{arrows.meta,positioning,fit}
\usepackage[numbers,sort&compress]{natbib}
\usepackage[hidelinks]{hyperref}
\usepackage[nameinlink,noabbrev]{cleveref}
\hypersetup{
  pdftitle={AstronOS: A Unified Execution Model and Runtime for Long-Horizon Agentic Systems},
  pdfauthor={Zhenhang Nie, Gui Zheng, Xudong Sun, Tailong Zhu, Bin Zhang},
  pdfsubject={Execution models and runtimes for long-horizon agentic systems},
  pdfkeywords={agentic systems, long-horizon execution, durable state, workflow runtime}
}

\setlist{nosep,leftmargin=*}
\newcolumntype{Y}{>{\raggedright\arraybackslash}X}
\newcolumntype{P}[1]{>{\raggedright\arraybackslash}p{#1}}
\newcommand{\system}{AstronOS}

\title{\textbf{AstronOS: A Unified Execution Model and Runtime\\for Long-Horizon Agentic Systems}}
\author{%
Zhenhang Nie\textsuperscript{1,\textdagger},
Gui Zheng\textsuperscript{1,\textdagger,*},
Xudong Sun\textsuperscript{1},
Tailong Zhu\textsuperscript{1},
Bin Zhang\textsuperscript{1,*}\\[0.45em]
\small\textsuperscript{1}iFLYTEK Co., Ltd., Hefei, Anhui, China.\\[-0.1em]
\small\textsuperscript{\textdagger}These authors contributed equally to this work.\\[-0.1em]
\small\textsuperscript{*}Corresponding authors: Gui Zheng and Bin Zhang.\\[-0.1em]
\small Contact emails:
\href{mailto:zhnie2@iflytek.com}{\nolinkurl{zhnie2@iflytek.com}};
\href{mailto:guizheng@iflytek.com}{\nolinkurl{guizheng@iflytek.com}}.
}
\date{}

\begin{document}
\maketitle

\begin{abstract}
Agentic systems often organize execution and state around a single conversation,
model invocation, or agent instance.  Once the conversation, model invocation,
or agent instance ends, the system may no longer retain a continuous
representation of the work, even though the real task can span many calls and
stages.  We introduce a unified execution model in which the system maintains
the work item's persistent identity and versioned authoritative state---the
currently accepted facts, decisions, and artifact references---across calls and
stages.  Each step receives scoped input constructed from a specific state
version together with the material newly available for that step.  A result
advances state only after validation passes and the state update is recorded
  successfully as a new version.  We implement this unified execution model on
  selected AstronOS runtime paths built around Cases, Tasks, and Scenario Packs
  across central orchestration and local execution.

To compare complete strategies for carrying an established plan into a fresh
model session, we design ten controlled tasks centered on planning a
software-version update.  Each task gives the model a set of candidate changes,
their dependency relations, and current test-status information.  The model
must decide which changes enter the update, which are deferred, and in what
order the selected changes should be handled.  Five tasks provide all material
at once and are completed in one model session.  The task materials and scoring
requirements are generated and fixed before execution.  The other five provide material
in three steps, each beginning in a fresh session.  A later session must carry
forward the previously agreed scope and adjust the plan when new test-status
information arrives.  This setup compares complete conditions for handing an
established plan to a later model session.

In the experiment, AstronOS carries the decision state through Cases, Tasks,
and Scenario Packs.  We compare five complete prior-state strategies: re-reading the
original task materials, replaying the full history, reading a deterministic
text summary, reading a deterministic JSON record, and using the AstronOS
runtime-mediated handoff.  Ten instances are run under all five strategies
with three repetitions, for 150 included executions.  On the single-stage reference
family, the strategies perform similarly.  In the primary A--C batch of
three-stage instances, AstronOS passes the frozen scorer in 14 of 15 executions,
compared with 0/15 for A and 2/15 for B.  D and E, collected in a later
non-interleaved batch, each pass 0/15.  Its attempt-accounted model-token cost
per passing execution is lower, while its condition-specific execution-window
time per attempt and time in the predominantly single-stage both-passing subset
are higher.  In this benchmark, condition C---its prior-state payload and
AstronOS execution path taken together---is associated with a higher end-to-end
scorer pass rate across fresh sessions, at a measurable time cost.
\end{abstract}

\section{Introduction}
\label{sec:introduction}

Language-model agents can interleave reasoning with actions in interactive
environments \citep{yao2023react}.  OpenHands exposes tools and a sandboxed
runtime for agents acting in software-development environments
\citep{wang2025openhands}.  Representative agent frameworks organize immediate
execution around conversations, message exchange, roles, or a task-local event stream
\citep{wu2024autogen,gao2024agentscope,hong2024metagpt,wang2025openhands}.
Such a unit is
convenient for a short interaction, but it is a poor stand-in for work that
proceeds through analysis, planning, implementation, and verification.  The
underlying work item persists even when the conversation ends, responsibility
changes hands, new evidence arrives, or a later step invalidates an earlier
assumption.

Keeping such work moving requires the system to maintain an up-to-date task
state for whoever takes the next step.  The next model or person must know which
facts and decisions are current, which artifacts they refer to, and which earlier
statements have been superseded.  If the system preserves no work-item state,
whoever takes over must re-read source
material and reconstruct that state.  If it replays every prior message and tool
record, the input grows with the execution history and leaves the model to infer
which parts still govern the task.  Long context does not by itself ensure that
models use all included information reliably \citep{liu2024lostMiddle}.
Multi-stage summarization can compress long dialogues and documents
\citep{zhang2022summn}, and memory systems can retain or retrieve prior
information \citep{packer2023memgpt}; however, neither mechanism alone specifies
which task state is authoritative or when a new output becomes committed state.

We focus on tasks that unfold over multiple steps.  Each step may establish a
new task state, and later steps must continue from that state rather than start
over.  We call these \emph{long-horizon agentic tasks}.  Here, long-horizon
refers to state dependence across steps, not simply long prompts or many model
turns.  The dependency may span sessions, models, people, tools, and execution
environments.

To support such tasks, we introduce a \emph{unified execution model}.  It gives
the overall task a persistent identity and a versioned authoritative state, and
specifies what information each step may receive and when a new result may
update task state.  An output from a model, person, or tool first enters as a
\emph{candidate result}.  Only after runtime checks and formal recording does
it become an \emph{accepted result} available to later steps.  A reported
benchmark pass additionally means that an independent scorer judges the
complete task execution correct.

\system{} provides a path-scoped implementation of the model.  For each work item that
must continue over time, the system creates a \emph{Case}: a unified record
that retains confirmed facts, decisions, artifacts, progress, and their
revisions.  From the current Case state, the system creates \emph{Tasks}.  Each
Task is a bounded unit of work that may be carried out by a model, a person, or
a tool.  A \emph{Scenario Pack} defines how a family of work is decomposed and
advanced.  As an executable
process template, it specifies the stages, which Tasks to create in each stage,
and the conditions a Task result must satisfy before the Case can advance.
Put differently, the Case retains the current state of the overall work, a
Task performs the current step, and the Scenario Pack determines what comes
next.

Before a Task runs, the system assembles the information needed for that step
from the Case and newly arrived material.  In the full product implementation,
this input is called a \emph{Context Bundle}.  It contains current facts,
decisions, relevant artifacts, and their provenance, so that the next executor
does not need to re-read the full history.  After a Task completes, the system
checks its result and records the accepted content back into the Case as the
basis for later steps.  Central services, the local workbench, models, and
tools can therefore coordinate around the same task state.

The paper proceeds in three layers.  The \emph{model layer} defines the unified
execution model and its operating rules.  The \emph{system layer} maps selected
\system{} runtime paths to the model and records their coverage boundaries.
The \emph{empirical layer} compares one complete \system{} condition with four
alternative prior-state strategies; it measures condition-level outcomes and
cost rather than runtime-wide conformance or individual component effects.

The evaluation uses ten controlled software-version update planning tasks.  Five tasks
provide all material at once as a single-stage reference family; five provide material over three steps and
require the agreed scope to be carried forward, revised after new test-status records,
and recorded as a final decision.  Each instance is run three times with the same
requested model identifier and reasoning setting, tools, stage-local facts, and
deterministic scorer
under five prior-state interfaces: re-investigation, full-history replay,
\system{} runtime-mediated state handoff, a rolling summary, and a lightweight JSON
record.  We treat each interface as a complete state-handoff strategy and compare
its task pass rate, token cost, and execution time.  Across the primary A--C
batch, \system{} passes the scorer in 93.3\%, compared with 50.0\% for A and
53.3\% for B; later-batch D and E pass 50.0\% and 46.7\%.  On the three-stage
family, the corresponding counts are C 14/15, A 0/15, B 2/15, and later-batch
D/E 0/15.  Its attempt-accounted model tokens
per observed pass are lower, while the predominantly single-stage both-passing
subset has higher condition-specific execution-window time.  Taken together, the result is a
completion--token--latency trade-off.

The paper makes three contributions:
\begin{itemize}
  \item We introduce a unified execution model for long-horizon agentic tasks.
  The model gives an overall task a persistent identity and versioned formal
  state across sessions and executors, and specifies how each step receives
  information, produces a result, and updates task state.
  \item We provide a path-scoped implementation of the model in \system{}.
  Cases retain work state, Tasks carry concrete steps, Scenario Packs drive
  progression, and Context Bundles prepare step-specific input.  A source and
  trace audit maps the model requirements to implemented mechanisms and records
  the current coverage boundaries.
  \item We compare five complete state-handoff strategies over 150 controlled
  executions.  In the primary A--C batch of three-stage tasks, \system{} passes
  14/15 executions, compared with 0/15 for A and 2/15 for B; later-batch D and E
  each pass 0/15.  Condition C has lower accounted model tokens per observed
  pass, with longer condition-specific execution-window time per attempt and in
  the predominantly single-stage both-passing subset.
\end{itemize}

\section{Related Work}
\label{sec:related}

\subsection{Agent runtimes and multi-agent frameworks}

Agent frameworks increasingly separate application logic from model and tool
execution.  AutoGen represents applications as conversations among customizable
agents that may combine models, tools, and people \citep{wu2024autogen};
AgentScope emphasizes message-centric composition, service integration, fault
tolerance, and distributed execution \citep{gao2024agentscope}; and MetaGPT
encodes standard operating procedures as structured multi-agent collaboration
\citep{hong2024metagpt}.  OpenHands provides an event-stream platform, sandboxed
runtime, tools, and evaluation interfaces for software agents
\citep{wang2025openhands}.  These systems make agent construction and execution
more systematic.  Across these systems, prominent abstractions include
conversations in AutoGen, messages and actors in AgentScope, roles and standard
operating procedures in MetaGPT, and event streams in OpenHands.
\system{} asks a complementary question: how to make the versioned work item and
its accepted state the primary continuity unit across heterogeneous executors.

AIOS is particularly relevant because it places scheduling, context, memory,
storage, tools, and access control behind an operating-system-style interface
\citep{mei2025aios}; Cerebrum provides associated development and distribution
facilities \citep{rama2025cerebrum}.  AIOS centers an OS-style kernel interface
for agent applications; our continuity unit is a versioned cross-step work item
and its accepted state, from which executions are derived.  The two systems
therefore model the execution boundary differently.

StateFlow represents task-solving control as states and transitions
\citep{wu2024stateflow}, while Agent Workflow Memory learns and retrieves
reusable procedural workflows across tasks \citep{wang2025awm}.  AstronOS
addresses a different continuity problem: maintaining the versioned,
authoritative state of one work item across fresh sessions and heterogeneous
executors.

\subsection{Durable workflows and case-centered processes}

Workflow research has long treated multi-step work as durable state rather than
as a transient process.  Case handling organizes flexible, knowledge-intensive
work around a case and its data \citep{vanderAalst2005caseHandling}.  Business
artifacts and Guard--Stage--Milestone lifecycles combine an information model
with event-driven lifecycle constraints \citep{hull2011gsm}, while Dynamic
Condition Response (DCR) graphs
provide declarative event constraints for adaptable processes
\citep{hildebrandt2011dcr}.  Durable workflow engines persist progress so that
long-running execution can survive failures \citep{burckhardt2022netherite};
Sagas provide a foundational decomposition and compensation model for long-lived
transactions \citep{garciaMolina1987sagas}.

The model's revision check also draws on optimistic concurrency control, which
allows tentative work to proceed and validates conflicts before commit
\citep{kung1981optimistic}.  The contribution lies in the model-facing
composition of versioning, compare-and-set, and validation:
the state version used to prepare an executor's input and the later
candidate-to-commit relation are both explicit and auditable.

Building on this lineage, we focus on a specific question: when work unfolds
over multiple steps, how can a system carry confirmed state into the next step
while ensuring that model output changes official state only after validation?
We summarize each step as Prepare--Execute--Check--Commit.  The system first
prepares Task input from the current Case state and its Scenario Pack.  A model,
person, or tool performs the Task and returns a candidate result.  The system
then checks whether that result still applies to the current Case version; only
after the check succeeds is the result written to the Case for later steps.
The Scenario Pack governs progression, the Case holds official state, and the
model, person, or tool performs the current step.

\subsection{Long context, summaries, and agent memory}

Several lines of work address continuity by extending or managing what a model
can see.  MemGPT applies a virtual-memory analogy to move information between
context and external storage \citep{packer2023memgpt}.  Generative Agents retain
an experience stream, retrieve memories, and synthesize reflections
\citep{park2023generativeAgents}.  Multi-stage summarization compresses inputs
that exceed a model's window \citep{zhang2022summn}.  Conversely, evidence that
models can underuse information in the middle of a long prompt cautions against
equating context capacity with effective state continuity
\citep{liu2024lostMiddle}.

The memory and summarization systems above primarily address how a model retains or retrieves past
information.  We further ask which state a later step should treat as official
when work unfolds over multiple steps.  AstronOS treats checked and recorded
Case content as the current official state and uses it to prepare the
information needed for the next step.  A later executor therefore receives the
work's currently confirmed state rather than only fragments of its history.

\subsection{Long-horizon agent evaluation}

Agent benchmarks increasingly use executable environments and final-state
checks.  SWE-bench evaluates whether language models can generate patches for
repository issues, verified by tests \citep{jimenez2024swebench}; WebArena evaluates functional task
completion in reproducible web environments \citep{zhou2024webarena};
$\tau$-bench combines user interaction, tool use, policy constraints, and final
database state \citep{yao2025taubench}; and OSWorld evaluates open-ended tasks
across real desktop applications with programmatic checks
\citep{xie2024osworld}.  These benchmarks exemplify outcome-oriented evaluation,
but they primarily test agent capability within an environment.

Long-term memory benchmarks provide a closer continuity comparison.  LoCoMo
evaluates question answering, event summarization, and dialogue generation over
very long multi-session conversations \citep{maharana2024locomo}; LongMemEval
evaluates extraction, multi-session and temporal reasoning, knowledge updates,
and abstention in sustained user--assistant histories
\citep{wu2025longmemeval}.  Our benchmark instead scores end-to-end task
completion under five ways of carrying a structured work decision across fresh
executor sessions.

We use the same task material, model, reasoning effort, tools, and scoring rules
to compare five strategies for carrying prior information into the next stage.
The strategies differ in both the information shown to the model and the steps
used to run them, so the experiment measures each strategy as a whole.  The task
material is prepared in advance, which keeps the reference answers fixed and
allows every run to be scored under the same rules.  The conclusions currently
apply to these controlled tasks; broader real-world use requires further
validation.

\section{A Unified Execution Model}
\label{sec:model}
\subsection{Scope and terminology}

The model targets work whose cross-step state cannot be represented reliably by
one transient execution context.  We call a work item \emph{long-horizon} when
it is completed through multiple steps and a later step depends on state
committed after an earlier one.  Duration is incidental: a three-step work item
completed in five minutes is in scope if its final step depends on a previously
accepted scope decision or versioned artifact reference; an hour-long batch of
independent questions is not.

We use three terms throughout the model.  A \emph{work item} is the enduring
unit of state, identified by $c$.  A \emph{step} is one state-dependent unit of
work for that item, represented by $t$.  An \emph{attempt}, indexed by $k$, is
one invocation of an executor for a step; technical retries may create multiple
attempts for the same step.  In \system{}, work items and steps are instantiated
as Cases and Tasks, respectively, although the concrete revision binding can
span an Analysis Run and a Task (\cref{sec:runtime}).

\paragraph{Running example.}
Suppose the second stage of a version-update task selects a precise set of
changes.  The runtime records that decision in the work item's accepted state.
A third stage starts in a fresh session, receives the earlier decision together
with a newly arrived test-status record, and returns a candidate final decision.
Runtime admission means that the stage result passed the applicable checks and
was recorded; benchmark correctness is decided later by the deterministic
scorer.  In the notation below, the recorded state is $\sigma_c^v$, the third
stage is step $t$, the new record is $n_t$, and the candidate decision is
$r_{t,k}$.  This example connects the model, the AstronOS runtime objects, and
the staged evaluation.

The model represents the evolving work item independently of any executor and
supports steps performed by a language model, person, deterministic program, or
tool.  Its efficiency objective is to retain reusable accepted work-item state
and derive a current, step-scoped view at each step.  Version binding and gated
advancement protect the integrity of that state; the efficiency benefit comes
from reducing repeated reconstruction and ever-growing history replay.

\subsection{Logical state and the four boundaries}

For a work item $c$, let $\sigma_c^v$ denote its authoritative state at logical
version $v$.  A version is an immutable, monotonically ordered snapshot in the
protocol model.  It contains accepted facts, decisions, artifact references,
and progress, plus lineage that identifies where those records came from.
For mutable external artifacts, the protocol stores a versioned reference or
content identifier.  ``Authoritative'' has a protocol-local meaning: later steps
use $\sigma_c^v$ as their current basis, without implying legal or organizational
authority.

We call the model's four properties \emph{boundaries} because each separates
objects that conversation-centered execution can conflate: the enduring work
item from its transient executor, accepted state from history and candidates, a
step-visible view from the full record, and an executor output from a committed
transition.  \Cref{tab:boundaries} states these separations.

\begin{table}[t]
  \centering
  \caption{The four boundaries of the unified execution model.}
  \label{tab:boundaries}
  \begin{tabularx}{\linewidth}{@{}p{0.20\linewidth}YY@{}}
    \toprule
    Boundary & Separation & Observable requirement \\
    \midrule
    Identity & enduring work item vs. session or executor & Every step and transition names exactly one work item. \\
    Authority & accepted state vs. history or candidate output & Later steps identify the accepted state on which they rely. \\
    Visibility & step-scoped view vs. full state or transcript & Prepared input identifies its state basis and new material. \\
    Commit & executor output vs. authoritative transition & Only a successfully committed change creates a successor state. \\
    \bottomrule
  \end{tabularx}
\end{table}

\subsection{Step record and execution protocol}

A step record is
\begin{equation}
  t=(\mathit{stepId},c,v_b,g_t,n_t,\pi_t),
  \label{eq:task}
\end{equation}
where $v_b$ is the base state version, $g_t$ is the step goal, and $n_t$ is
explicitly identified material that became available for this step but is not in
the base snapshot.  The policy $\pi_t$ specifies context preparation,
acceptance, and update rules.  The step identity is distinct from $c$: many
steps may contribute transitions to the same work item.

One attempt follows Prepare--Execute--Check--Commit:
\begin{align}
  b_t &= \textsc{Prepare}_{\pi_t}(t,\sigma_c^{v_b},n_t), \\
  r_{t,k} &= \textsc{Execute}(t,k,b_t), \\
  (d_{t,k},\Delta_{t,k})
    &= \textsc{Check}_{\pi_t}(t,\sigma_c^{v_{\mathrm{cur}}},r_{t,k}), \\
  (\sigma',o_{t,k})
    &= \textsc{Commit}_c(v_{\mathrm{cur}},t,k,d_{t,k},\Delta_{t,k}).
  \label{eq:cycle}
\end{align}
Here $b_t$ is the step-visible input, $r_{t,k}$ is a \emph{candidate result},
$d_{t,k}\in\{\mathsf{pass},\mathsf{reject}\}$ is the check decision, and
$\Delta_{t,k}$ is a proposed state change when the decision passes.  Commit is
atomic with respect to the protocol state of one work item.  It returns
$o_{t,k}=\mathsf{accepted}$ and $\sigma'=\sigma_c^{v_{\mathrm{cur}}+1}$ only if
the declared update policy remains valid at commit time and no earlier attempt
of the same logical step has already committed.  Otherwise it returns a
non-accepting outcome and leaves $\sigma'=\sigma_c^{v_{\mathrm{cur}}}$.

An executor's output therefore becomes an \emph{accepted result} only when its
derived change commits successfully.  A check that passes but loses a version
race is not acceptance.  This formulation closes the check-to-commit interval:
freshness is revalidated by Commit rather than assumed from an earlier read.
Atomicity here covers the protocol's authoritative state, not arbitrary external
actions performed by a tool.

If $v_b\neq v_{\mathrm{cur}}$, Check must reject unless $\pi_t$ declares an
explicit rebase operation.  A rebase reconstructs the proposed change against
$\sigma_c^{v_{\mathrm{cur}}}$ and reruns every acceptance predicate.  Commit
names the version that was checked and succeeds only while that version remains
current.  Merely checking a stale candidate against unseen current state is not
a valid rebase.

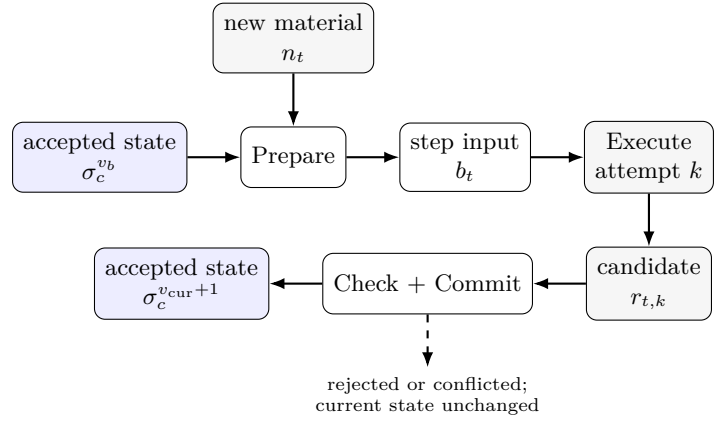
\begin{figure}[t]
  \centering
  \begin{tikzpicture}[
    node distance=7mm and 7mm,
    box/.style={draw,rounded corners,align=center,minimum height=8mm,inner sep=4pt,font=\small},
    state/.style={box,fill=blue!7},
    outside/.style={box,fill=gray!7},
    flow/.style={-{Latex[length=2mm]},thick},
    reject/.style={-{Latex[length=2mm]},thick,dashed}
  ]
    \node[state] (base) {accepted state\\$\sigma_c^{v_b}$};
    \node[box,right=of base] (prepare) {Prepare};
    \node[box,right=of prepare] (input) {step input\\$b_t$};
    \node[outside,right=of input] (execute) {Execute\\attempt $k$};
    \node[outside,below=of execute] (candidate) {candidate\\$r_{t,k}$};
    \node[box,left=of candidate] (commit) {Check + Commit};
    \node[state,left=of commit] (next) {accepted state\\$\sigma_c^{v_{\mathrm{cur}}+1}$};
    \node[outside,above=of prepare] (new) {new material\\$n_t$};
    \draw[flow] (base) -- (prepare);
    \draw[flow] (new) -- (prepare);
    \draw[flow] (prepare) -- (input);
    \draw[flow] (input) -- (execute);
    \draw[flow] (execute) -- (candidate);
    \draw[flow] (candidate) -- (commit);
    \draw[flow] (commit) -- (next);
    \draw[reject] (commit.south) -- ++(0,-7mm) node[below,font=\scriptsize,align=center]{rejected or conflicted;\\current state unchanged};
  \end{tikzpicture}
  \caption{A step binds input to accepted state while keeping candidate output
  outside that state until Check and Commit succeed.}
  \label{fig:execution-cycle}
\end{figure}

\paragraph{Prepare is policy-parameterized.}
The protocol requires $b_t$ to identify its base state and the lineage of
state-derived and newly supplied factual records.  Minimality, completeness for
a particular executor, and semantic optimality depend on $\pi_t$ and the task
family and must be evaluated empirically.  The experiment therefore compares
complete prior-state interfaces; a semantic-equivalence control that changes
only surface format remains future work.

\subsection{Trace-level design obligations}

The four boundaries induce five obligations that a runtime can test on an
execution trace:

\begin{description}[style=nextline]
  \item[O1 --- Identity isolation.] A transition belongs to exactly one $c$ and
  does not modify the authoritative state of another work item.
  \item[O2 --- Snapshot binding.] A step identifies its base version, and its
  prepared input records the state basis actually used.
  \item[O3 --- Candidate isolation.] Producing or storing $r_{t,k}$ does not by
  itself modify protocol-level authoritative state.
  \item[O4 --- Lineage coverage.] Every factual record copied from state or new
  material into $b_t$ identifies a source kind and source identifier or version.
  \item[O5 --- Commit isolation.] A successful Commit uniquely links the step,
  accepted attempt, change, and successor version; failed or rejected attempts
  create no successor version.
\end{description}

These obligations cover protocol-level lineage and commit isolation.  The truth
of accepted facts, the soundness of an acceptance rule, optimal context
selection, and transactional isolation of arbitrary external tool effects lie
outside that scope.  Section~\ref{sec:runtime} maps each obligation to mechanisms
in a pinned \system{} revision and reports observations on audited paths;
runtime-wide conformance would require broader testing.

\subsection{Failure, retry, and concurrency}

Failures are classified by where an attempt stops.  If Execute fails, no
candidate is available and protocol state remains unchanged.  If Check rejects a
candidate, the runtime may record the attempt and create a corrective step, but
the rejected content is not committed.  A transient provider or network event
may trigger a technical retry under a declared policy; the retry remains an
attempt of the same step unless the runtime explicitly creates a replacement
step.

Concurrent or delayed results are handled at Commit.  Work need not execute
serially, but a proposed change must still be valid against the current version.
If it is not, the runtime leaves authoritative state unchanged and may prepare a
new step from the current state.  External tools may already have acted before
returning a candidate; such effects require idempotency, compensation, or a
recoverable workflow outside the atomic protocol-state commit.  Concurrency and
recovery support the reusable state boundary but are not the empirical focus of
this paper.

\subsection{Observable evaluation expectations}
\label{sec:hypotheses}

Preparing and carrying a step-scoped view incurs additional per-step work: the
runtime loads state, selects records, preserves lineage, and coordinates
execution.  It is useful only if it avoids greater loss from repeated
investigation, history ambiguity, or unsuccessful executions.  We state three
condition-level expectations that match the available measurements:

\begin{itemize}
  \item \textbf{E1 (scorer pass rate).} With the requested model identifier and
  reasoning setting, tools, stage-local evidence, and scorer fixed, complete
  condition C---including its prior-state payload and \system{} runtime
  path---should exhibit a higher observed fraction of executions that pass the deterministic scorer on tasks
  with cross-step dependencies.
  \item \textbf{E2 (token efficiency).} Under a fixed attempt budget, the
  condition should consume fewer total observable model tokens per passing
  execution.  This ratio is a benchmark-level cost-effectiveness measure, not an
  estimate of a sequential retry-until-success process and not a claim that each
  prompt is shorter.
  \item \textbf{E3 (end-to-end latency).} Additional runtime work may increase
  latency in both-passing comparisons, especially for short or
  single-stage tasks.  The experiment measures the net difference but cannot
  attribute it to individual runtime components.
\end{itemize}

These are condition-level expectations.  Attributing any observed difference
to versioning, lineage, validation, product Context Bundles, or another
protocol component requires separate controls (\cref{sec:method}).

\subsection{Non-goals}

The unified execution model complements language models, memory retrieval
algorithms, and workflow notations.  Case-centered process management,
artifact lifecycles, and durable workflows provide precedents for case-oriented
state, data-aware lifecycle control, and persistent workflow execution
\citep{vanderAalst2005caseHandling,hull2011gsm,burckhardt2022netherite}.  Case
persistence, workflow stages, and optimistic concurrency are foundations,
not separate novelty claims of this paper.  Our contribution is their
model-facing composition into a trace-checkable contract: each step names its
accepted-state basis; state-derived and newly supplied facts retain lineage;
and executor output remains a candidate until validation and a
version-conditioned commit create a successor state.  O1--O5 operationalize
this joint contract.  We then provide a path-scoped AstronOS implementation and
evaluate one bundled end-to-end strategy.

\section{The AstronOS Runtime}
\label{sec:runtime}
\subsection{Runtime scope and architecture}

\system{} is a distributed runtime placed between model executors and the
systems in which work is recorded or performed.  The pinned implementation used
in this paper is source revision
\path{0b0349d5290ce6d0949c6e57b646c053d437e2b6}.  Its central services retain the
Case record and advance Scenario-defined stages; a runtime manager invokes model
executors; and a local workbench executes Tasks that require access to a
developer's files, repositories, or tools.  The separation allows state to
remain central while execution occurs where the relevant resources are
available.

A Scenario Pack is the reusable process definition; when a Case is bound to one,
the runtime creates the corresponding per-Case Scenario.  We use \emph{Scenario
Pack} for the reusable definition and \emph{Scenario} for that runtime binding.

\begin{figure}[t]
  \centering
  \begin{tikzpicture}[
    node distance=8mm and 6mm,
    box/.style={draw,rounded corners,align=center,minimum height=9mm,
      minimum width=24mm,inner sep=3pt,font=\scriptsize},
    store/.style={box,fill=blue!7},
    exec/.style={box,fill=green!7},
    flow/.style={-{Latex[length=2mm]},thick},
    back/.style={-{Latex[length=2mm]},thick,dashed}
  ]
    \node[store] (case) {Hub + store\\Case snapshot $v$};
    \node[box,right=of case] (workflow) {Workflow Worker\\Scenario stage};
    \node[box,right=of workflow] (bundle) {Input assembly\\Context policy};
    \node[exec,below left=8mm and 4mm of bundle] (runtime) {Runtime Manager\\model executor};
    \node[exec,below right=8mm and 4mm of bundle] (edge) {Desktop Connector\\local tools};
    \node[box,below=24mm of bundle] (candidate) {candidate result};
    \node[box,left=of candidate] (check) {Check +\\protocol-state commit};
    \node[store,below=24mm of case] (next) {Case snapshot $v{+}1$};
    \draw[flow] (case) -- (workflow);
    \draw[flow] (workflow) -- (bundle);
    \draw[flow] (bundle) -- (runtime);
    \draw[flow] (bundle) -- (edge);
    \draw[back] (runtime) -- node[left,font=\scriptsize]{model output} (candidate);
    \draw[back] (edge) -- node[right,font=\scriptsize]{tool result + evidence} (candidate);
    \draw[flow] (candidate) -- (check);
    \draw[flow] (check) -- (next);
    \draw[flow] (next) -- node[left,font=\scriptsize]{next stage} (case);
  \end{tikzpicture}
  \caption{The \system{} path relevant to the execution model.  Case state is
  central; model and tool execution can be central or local.}
  \label{fig:runtime-architecture}
\end{figure}
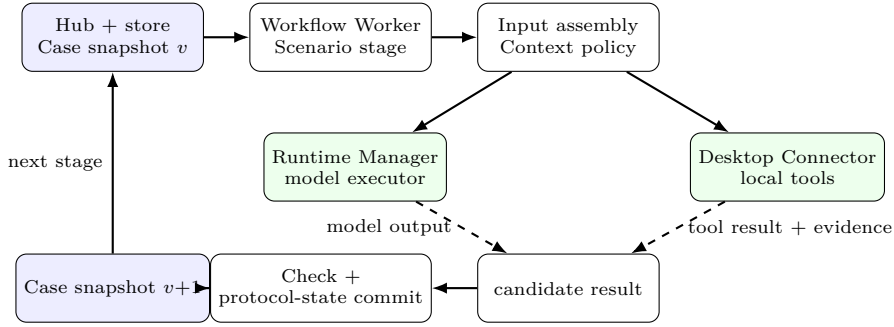

\Cref{fig:runtime-architecture} shows the data path relevant to this paper.  The
Hub API exposes Case and Task operations and persists Case snapshots.  The
Workflow Worker interprets Scenario Pack stages and schedules work.  The Runtime
Manager starts central model runs.  The Desktop Connector receives an edge Task,
executes it in the local workbench, and returns a typed result with artifacts or
structured data.  User-interface components are omitted because they do not
participate in the measured context mechanism.

\subsection{Mapping model objects to the implementation}

The formal objects in \cref{sec:model} span more than one database row.  In the
audited central-analysis path, a Case plus its latest Case Snapshot represents
the persistent work item and authoritative state.  An Analysis Run is created
from a specific snapshot revision.  A dispatched edge Task always names its
Case and may name the upstream analysis run and Context Bundle in its prompt
input.  The frozen implementation has no universal \texttt{Task.baseVersion}
field; explicit revision binding is complete on the Analysis Run path.  This
mapping is summarized in \cref{tab:runtime-mapping}.

\begin{table}[t]
  \centering
  \caption{Execution-model objects and their \system{} representation.}
  \label{tab:runtime-mapping}
  \begin{tabularx}{\linewidth}{@{}p{0.20\linewidth}p{0.28\linewidth}Y@{}}
    \toprule
    Model object & Runtime representation & Observable fields \\
    \midrule
    Work item & Case and Case Snapshot & Case identifier; status; workflow state; snapshot revision \\
    Step & Analysis Run and, when needed, a dispatched Task & Case identifier; run/Task identifiers; input revision on Analysis Runs; goal and expected result \\
    Step input & Product Context Bundle or the experiment-specific step input, plus stage-local material & objective; prior accepted result; current material; optional bundle sources \\
    Candidate & Run or Task result before interpretation & status; text/JSON body; artifact metadata; run/Task identifier \\
    Accepted change & Interpreted stage result merged into the snapshot & new snapshot revision and a recorded stage-result event \\
    \bottomrule
  \end{tabularx}
\end{table}

This distinction prevents a naming shortcut from hiding the actual
implementation.  The abstract Task in \cref{eq:task} corresponds to a complete
step execution; it is not necessarily identical to the row named
\texttt{TaskSchema}.  Here, a \emph{fence} is a deterministic guard that prevents
a result from advancing the Case unless it belongs to the current run or leased
Task.  For the central path examined here, the Analysis Run
carries the explicit input snapshot revision.  Edge Tasks retain Case identity
and current-Task fencing.  In the frozen implementation, explicit base-version
binding is limited to the audited Analysis Run path.

\subsection{Cases and versioned snapshots}

A Case stores stable identity and coarse workflow status.  Its latest Case
Snapshot stores the current title, summary, structured state, entities, and
evidence under an integer revision.  Snapshot updates increment the revision.
The repository uses a compare-and-set update on the revision and retries a merge
if another writer advanced the Case first.  This compare-and-set update gives
the state transition a concrete concurrency boundary and avoids
last-write-wins behavior.

This repository retry re-applies a merge function to the newest row and serves
as a storage-level conflict mechanism.  The explicit semantic rebase in
\cref{sec:model} additionally requires task-level recomputation.  The audited central-analysis path first ignores a
result whose input revision is already stale.  Our O5 correspondence therefore
covers the observed no-conflict commit path and atomic latest-row advancement.
Concurrent merges may proceed without rerunning every task-level acceptance
predicate.

The snapshot table stores only the latest revision.  Analysis requests archive
the input snapshot they send, and events record state transitions; together they
provide execution evidence rather than a fully addressable database history of
immutable snapshots.  In this paper,
``versioned state'' means an incremented latest-state revision with optimistic
concurrency and archived execution evidence.

Central analysis is tied to that boundary.  The runtime constructs an execution
identifier from the Case identifier and snapshot revision, sends the full input
snapshot and its revision to the Runtime Manager, and stores the same revision
on the Analysis Run.  When the result returns, the outcome activity reloads the
latest snapshot.  If the revision no longer matches, it records that the outcome
was not interpreted; otherwise it interprets the stage result and merges the
accepted patch into a newly revised snapshot.  This is the exercised
implementation of version binding and candidate isolation.

\subsection{Context Bundle construction}

A Context Bundle is a typed record associated with a Case, stage or run.  Its
schema separates the step objective, facts, decisions, constraints, artifact
references, optional memory references, source records, and information that
must not be included.  Each product source can retain its type, object identity,
summary, sensitivity, inclusion or redaction status, and reason for selection.

Before starting a central analysis run, the workflow path resolves or creates
the Case's current bundle, hydrates a model-facing summary, and attaches the
bundle identifier and summary to the run request.  It records both bundle
creation and bundle use as Case events, including the run identifier and source
count.  The content seen by the model is therefore linked to a stored,
traceable bundle.  This product path is established
by contracts, repositories, dispatch code, and tests in the pinned revision.

\subsection{The runtime-mediated state-handoff interface used in the experiment}
\label{sec:experiment-interface}

Condition C uses the \system{} Case/Task/Scenario runtime.  At each stage, the model receives a
hash-pinned package containing current-stage synthetic evidence.  When prior
state is needed, the runtime adds the runtime-admitted previous-stage result and a
structured stage-dispatch or decision artifact; later stages also receive the
record showing that the intermediate deterministic gate was satisfied.  These
objects are archived with content hashes and linked to the same Case and its
successive Tasks.

We call this condition the \emph{\system{} runtime-mediated state handoff}.  The name is
intentionally broader than a representation format: the condition uses the
runtime to carry committed stage state and to progress the same Case.  The
product Context Bundle and long-term-memory paths are outside this measured
condition.  The evidence therefore applies to the complete runtime-mediated
interface described above, not to either of those product paths.

\subsection{Scenario-driven progression and local execution}

Scenario Packs keep task-family semantics in configurable definitions rather
than hard-coding them in the generic runtime implementation.  A pack
declares stages, expected outputs, and progression conditions.  The Workflow
Worker reads the bound pack when deciding which analysis or edge Task to create.
The bound pack supplies workflow-specific semantics to the shared Case,
snapshot, Task, and result-handling machinery, allowing the Hub and connector to
remain domain-independent.  Product Context Bundles can be attached by the
context path, while Scenario progression is configured independently.

When a product stage requires repository or local-tool access, the Hub dispatches
a typed Task to the Desktop Connector.  The Task includes the Case identifier,
execution target, operator prompt, expected result mode, timeout, idempotency
key, and optional context and skill references.  The connector returns a typed
Task result containing status, a summary, optional text or JSON, artifacts, and
session information.  This describes a product path; the benchmark in this
paper uses synthetic tools and does not access a repository.  Returning success
at this transport boundary is not by itself a Case-state update: the central
workflow still interprets the result against the applicable stage conditions.

\subsection{Implementation conformance evidence}
\label{sec:conformance-trace}

We checked the pinned contracts, repositories, runtime activities, result
pipeline, and corresponding tests against the model obligations.  The code
contains path-specific mechanisms corresponding to all five: Case and Task
identifiers support identity linkage (O1); the central-analysis path records and
fences on its input snapshot revision (O2); a returned Task result remains
non-authoritative until it passes current-Task, dispatch-lease, and Scenario
interpretation fences (O3); product Context Bundles expose source records and
creation/use events that can support lineage (O4); and a revision-conditioned
snapshot merge supports latest-state advancement on the audited no-conflict
path (O5).  These observations establish implementation correspondences on the
inspected paths.  Runtime-wide satisfaction of O1--O5 requires broader testing.

Three implementation boundaries qualify the correspondence.  Edge Tasks lack a
mandatory base-version field; only the Analysis Run path has explicit snapshot
binding.  Candidate is a protocol role played by a stored run or Task result.
Finally, snapshot merge is atomic at the Case row, while result persistence,
workflow signaling, and external actions form a fenced, recoverable pipeline
with no cross-system transaction.  These boundaries come from implementation
inspection.

\begin{table}[t]
  \centering
  \caption{Scope of the implementation and benchmark evidence for O1--O5.}
  \label{tab:obligation-coverage}
  \scriptsize
  \begin{tabularx}{\linewidth}{@{}p{0.08\linewidth}p{0.38\linewidth}Y@{}}
    \toprule
    Obligation & Pinned implementation evidence & Exercised in condition C \\
    \midrule
    O1 & Case/Task identity linkage & Observed in the audited trace \\
    O2 & Analysis Run snapshot-revision binding & Not established for the C Task path; no universal Task base version \\
    O3 & Current-Task, lease, and Scenario result fences & Observed on the selected path \\
    O4 & Product Context Bundle source lineage & Implemented, but not exercised by this benchmark \\
    O5 & Revision-conditioned latest-row merge & Audited on a no-conflict path; concurrency/rebase not tested \\
    \bottomrule
  \end{tabularx}
\end{table}

\subsection{An audited experiment trace}

\Cref{tab:trace} presents a sanitized successful C execution from the formal
evidence archive.  One Case persists across four Tasks: three synthetic
decision stages and one checkpoint-propagation Task.  The trace is an execution
of the implemented AstronOS runtime path using synthetic domain material; it demonstrates
runtime-mediated state handoff in the measured path.

\begin{table}[t]
  \centering
  \caption{Sanitized trace C-1: one successful three-stage execution.}
  \label{tab:trace}
  \begin{tabularx}{\linewidth}{@{}p{0.13\linewidth}Y@{}}
    \toprule
    Relative time & Audited event \\
    \midrule
    0.0 s & Create Case C1 and snapshot r1. \\
    1.5--1.8 s & Create and dispatch T1 (stage 1); record the transport-level Task acknowledgement and archive hashes for the current-stage input and continuity payload. \\
    18.8 s & T1 returns a candidate result; the terminal pipeline interprets and admits it; snapshot advances to r2. \\
    19.6--19.8 s & Dispatch T2 (stage 2) with T1's runtime-admitted result and the structured decision artifact. \\
    33.6--34.1 s & T2 returns; snapshot advances through r3 and the deterministic checkpoint to r4. \\
    34.9--35.1 s & Dispatch T3, the checkpoint-propagation Task, with T2's result and the checkpoint receipt. \\
    62.7--62.8 s & T3 completes; snapshot advances to r5. \\
    63.5--63.8 s & Dispatch T4 (stage 3) with the preceding accepted state. \\
    81.5--81.6 s & T4 completes; snapshot advances to r6; Case C1 becomes resolved. \\
    Postflight & The independent deterministic scorer assigns 100/100 and marks the execution correct. \\
    \bottomrule
  \end{tabularx}
\end{table}

Trace C-1 corresponds to cell \texttt{GAR-101-r1-C}.  The manifest maps that
cell to an internal raw execution record; that record contains the full Case
timeline, model-visible prompt archive, continuity payload, and postflight
score.  The compact public artifact does not include this raw record.  The
trace records an execution of the implemented AstronOS Case/Task/Scenario path while its domain
materials and tool effects remain synthetic or simulated.  It connects the
model, implementation, and measured condition on an internally audited example;
exhaustive runtime conformance requires broader coverage.

\subsection{System claim boundary}

The implementation evidence establishes model correspondence on the audited
path.  Runtime-wide conformance and efficiency are separate claims.  The
experiment measures the complete prior-state interface defined above;
component-level effects of snapshot versioning, Scenario
progression, and result interpretation require additional controls.  Keeping
these claims separate is essential: source and trace evidence
answer whether the product implementation exposes the corresponding runtime paths, whereas the experiment asks
whether one end-to-end state-handoff strategy is useful under the frozen task
set.

\section{Evaluation Methodology}
\label{sec:method}
\subsection{Research questions}

The experiment compares the complete prior-state provision conditions described in
\cref{sec:experiment-interface}.  It asks three questions aligned with the
expectations in \cref{sec:hypotheses}:

\begin{description}[style=nextline]
  \item[RQ1 --- Scorer pass rate.] With task facts, requested model identifier
  and reasoning setting, tools, and scorer fixed, how often does each condition
  pass the deterministic task checks,
  particularly when a later stage must carry forward and revise prior state?
  \item[RQ2 --- Accounted token efficiency.] How many provider-reported model
  tokens are observed per passing execution, and how do token counts differ when
  both paired executions pass?
  \item[RQ3 --- Time trade-off.] How much condition-specific execution-window time and archived harness time is spent
  per attempt and per observed pass, and what difference remains among pairs in
  which both conditions pass?
\end{description}

The experiment's unit of comparison is the complete condition.  Its inference
scope is therefore limited to condition-level differences.

\subsection{Controlled staged version-update planning tasks}

The suite contains ten parameterized tasks generated from two templates, five
tasks per template.  In each task, the model prepares a plan for a software-version
update: it reads candidate changes, dependency relations, and test-status records
generated before execution; chooses which changes to include and which to hold;
and orders the selected changes while respecting their dependencies.  Task materials
are generated before the experiment, and model decisions are recorded only in an
in-memory simulator.  The suite therefore evaluates whether a staged plan can be
carried forward and revised accurately, rather than a complete software-development
workflow.

The \textbf{all-material-at-once family (S)} is a single-stage reference family.  Each instance provides six
change records with dependencies and pre-generated test-status records.  The
executor must compute
the exact releasable set, hold the failed optional leaf, produce a dependency-
valid order, and submit one structured decision.  S serves as a single-stage
reference because all relevant facts arrive in one stage.  S and GAR use
different generators, task sizes, and evidence structures, so their family
contrast is descriptive rather than a matched staging intervention.

We use \textbf{GAR} as a short label for the \textbf{material-in-three-steps family}.
Each instance contains eight changes, a
dependency graph, nine test identifiers, and three task stages.  Stage 1
identifies a failed optional leaf and records which evidence remains pending.
Stage 2 receives new pre-generated test-status records, computes the exact candidate scope,
and records a deterministic checkpoint tied to that scope.  In C, the runtime then
passes the runtime-admitted decision and checkpoint receipt through one
additional model-using \emph{checkpoint-propagation Task} before stage 3.  Its
benchmark role is to carry those records forward without changing the simulated
task state; its token and time costs are included.
Stage 3 receives the final simulated test-status update and must reuse the
runtime-admitted scope to record the final simulated release decision.  The
checkpoint-propagation Task has no external organizational effect in the simulator; it makes an exact intermediate decision necessary after
new evidence arrives.

Each family uses one shared generating template with five parameterizations.
Three stochastic repetitions per instance measure within-instance model
variability.

\subsection{Prior-state conditions}

All conditions use the same generated instance, current-stage material,
requested model identifier and reasoning setting, allowed synthetic tools,
output schema, and scorer.  The
condition-specific timeout policies are frozen and reported below.
Each stage starts a fresh provider session, so vendor-side conversation state is
not a hidden continuity channel.  The comparison varies the complete interface
through which prior-stage information becomes available
(\cref{tab:conditions}); as detailed below, that interface also changes runner
structure and information packaging.
Conditions A, B, D, and E use one common lightweight direct runner; condition C
uses the implemented AstronOS Case/Task/Scenario runtime path.

\begin{table}[t]
  \centering
  \caption{Five prior-state provision conditions.}
  \label{tab:conditions}
  \begin{tabularx}{\linewidth}{@{}P{0.08\linewidth}P{0.22\linewidth}Y@{}}
    \toprule
    ID & Condition & Information available at a later stage \\
    \midrule
    A & Re-investigation & Current-stage material; the executor may re-read the same synthetic sources with the same tools when it needs earlier information. \\
    B & Full history & Current-stage material plus all previous visible prompts, model outputs, and tool observations in their original order. \\
    C & \system{} runtime-mediated state handoff & A hash-pinned current-stage package plus the runtime-admitted prior-stage result, structured decision artifact, and applicable checkpoint receipt, materialized through the implemented Case/Task/Scenario runtime path. \\
    D & Rolling summary & Current-stage material plus a deterministic prose summary generated from the latest runtime-admitted decision by a frozen template. \\
    E & Lightweight JSON & Current-stage material plus a deterministic fixed-field JSON record generated from the latest runtime-admitted decision by a frozen function. \\
    \bottomrule
  \end{tabularx}
\end{table}

Conditions D and E are deterministic projections of the preceding
runtime-admitted decision, avoiding a second model and an additional source of
stochastic error.  They preserve the decision's fact
references, candidate, release and held sets, pending test-status items, execution order,
control-step fields, and next action.  D renders these fields as frozen prose; E
retains them as fixed-field JSON.

Condition C evaluates a runtime-mediated handoff on the Case/Task/Scenario path,
with product Context Bundle construction, memory extraction, and retrieval
disabled.  The treatment is the
complete prompt-and-orchestration stack: C also differs from the direct
conditions in stage packaging, the additional checkpoint-propagation Task, runtime
progression, and immediate information coverage.  Exact input text and
orchestration therefore vary together; attributing an effect to prose versus
JSON, versioning, or another component requires additional factorial controls.

\subsection{Frozen configuration and run matrix}

The task templates, prompts, condition builders, scorer and exclusion rules,
model configuration, and evidence schema were fixed in source revision
\path{0b0349d5290ce6d0949c6e57b646c053d437e2b6}, committed on 2026-08-09 at
14:52 China Standard Time, before the first included execution at 15:22.
Earlier pilot runs informed the harness design; no task family was held out or
preregistered, so the results are development-suite evidence.  The workbench used
Codex CLI 0.145.0 and requested \texttt{gpt-5.6-sol} with reasoning effort
\texttt{medium}; the archived executions report the same model identifier.
Memory extraction, vector retrieval, and long-term memory are disabled in every
condition.  The direct runner places a 15-minute provider-turn wait inside a
30-minute Task envelope; C places a 14-minute whole-Case deadline inside a
15-minute Task envelope.  No included execution reached these limits.  The
runner and provider evidence do not expose a
temperature, sampling seed, provider snapshot, context-window setting, or
maximum-output setting, so we do not claim to have pinned those controls.  No
included execution ended with a recorded context-length error.

Each of 10 instances is run under five conditions and three repetition indices,
yielding $10\times5\times3=150$ valid executions, 30 per condition.  Conditions
A--C were collected in the primary batch; D and E were collected in a
subsequent batch with the same source revision, generated tasks, requested model
identifier and reasoning setting, workbench, and evidence fields.  The provider
snapshot, seed, and sampling controls were not exposed.  The batches were
sequential rather than fully interleaved.  Within the primary batch, the six
possible A/B/C orders each appeared in five execution blocks; the supplemental
batch used a fixed D-then-E order.  Cross-batch contrasts in pass rate,
token use, and time are descriptive and may include provider drift as well as
the intended strategy differences.

Provider-capacity attempts were invalidated before inclusion and retried under
the same instance, condition, and repetition identifier.  Such attempts are
absent from the 150 valid executions and from the token/time accounting below.
The internal retry archive records three such physical attempts, one each under
A, D, and E; the compact manifest encodes only the two D/E exclusions.  A
provider-capacity event is therefore a technical retry, whereas a completed
execution that fails the scorer remains a task failure.

\subsection{Deterministic scoring checks}

The local TypeScript scorer consumes the structured decisions and simulated
tool events after execution; it does not call a language model.  Passing is
binary and stricter than reaching a diagnostic-score threshold.  S requires an
exact release/hold classification, dependency-valid canonical order, correct
failed-test-status evidence, a complete JSON decision, and exactly one simulated final
submission.  GAR additionally requires the correct state at all three stages,
the exact stage-2 scope and checkpoint, reuse of that scope after the final
evidence update, complete structured outputs, and one final simulated
submission.  Any missing critical element makes the execution non-passing even
if other scored items are correct.

The separate 100-point score is used only to locate partial failures.  For S,
the weights are 40 points for release/hold classification, 25 for dependency
order, 20 for test-status evidence, and 15 for output completeness.  For GAR,
the weights are 25 for final classification, 15 for dependency order, 15 for
test-status evidence, 20 for three-stage progression, 15 for checkpoint and
control handling, and 10 for output completeness.  These weights do not relax
the binary rule: a run can score 90 and still fail if a required stage, fact,
order, checkpoint, or final simulator event is missing or incorrect.

We use \emph{passing execution} or \emph{correct completion} only for an
execution that satisfies these deterministic checks.  This scorer decision is
the research outcome; \texttt{SUCCEEDED} records runtime status.

The model's term \emph{accepted result} means that a candidate was committed to
protocol state, independently of factual correctness.  To keep that distinction
visible here, we call an intermediate output that is persisted or
passed to the next stage \emph{runtime-admitted}.  The hidden deterministic
scorer subsequently determines whether each stage is postflight-correct and
whether the full execution passes.

\subsection{Metrics and accounting boundary}

For condition $q\in\{A,B,C,D,E\}$, let $N_q$ be the number of included
executions ($N_q=30$ for a condition-level summary and $N_q=15$ within one task
family).  For execution $i$, let $y_i\in\{0,1\}$ denote the deterministic
scorer outcome, $z_i$ accounted model tokens, $\ell_i$ condition-specific execution-window time,
and $h_i$ the archived harness-time field.  We report
\begin{align}
  \widehat{p}_q &= \frac{1}{N_q}\sum_i y_i, \\
  Z_q^{\mathrm{pass}} &= \frac{\sum_i z_i}{\sum_i y_i}, \\
  L_q^{\mathrm{pass}} &= \frac{\sum_i \ell_i}{\sum_i y_i}, \qquad
  H_q^{\mathrm{pass}} = \frac{\sum_i h_i}{\sum_i y_i}.
\end{align}
The latter three are \emph{attempt-accounted costs per observed pass}: their
numerators pool every valid attempt, including failures, and their denominators
are the observed passes.  Conditional means among passing executions and
estimates for a sequential retry-until-success policy are different quantities.  We also
report the ordinary per-attempt means $\sum_i z_i/N_q$,
$\sum_i\ell_i/N_q$, and $\sum_i h_i/N_q$.

Total tokens are accounted provider-reported input plus output tokens in the
experiment ledger.  For direct conditions, the ledger sums all observed
workbench turns, including same-stage recovery turns.  For C, it includes
archived domain execution, the checkpoint-propagation Task, and recorded model-using
control-plane work.  Stage-package text consumed by a model is included in
input tokens.  Deterministic state materialization, local scoring, database,
network, and CPU work are not tokenized; complete platform-wide control-plane
telemetry is unavailable.  We therefore call this \emph{accounted model-token
usage}; it covers the recorded model-usage boundary.

For A, B, D, and E, execution-window time starts immediately before the first task stage
and ends when the final simulated decision record has formed, before local
scoring and evidence assembly.  For C, it starts immediately before Case
creation and ends when the terminal Case or terminal Task failure is observed.
The archived harness-time field is condition-specific rather than one common
timer.  For A, B, D, and E, it is recorded inside the direct runner after the
outer Workbench preflight and before final evidence assembly.  For C, it is the
live Case-runner timer and includes that runner's service preflight, archival
work, and cleanup; postflight scoring occurs after the timer stops.  We report
this field as an auxiliary end-to-end ledger measure and do not interpret its
difference from execution-window time as a decomposed platform overhead.

For each instance--repetition pair in which C and baseline $b$ both pass, we
compute $(x_C-x_b)/x_b$ and report the median for tokens, execution-window time, and
the archived harness-time field.  This conditions on both observed outcomes and yields a
descriptive comparison of the both-passing subset.  Causal efficiency requires
a different design.  Pass rate, token cost, and time are the primary endpoints;
read counts and rereads used to recover earlier information are diagnostics.

\subsection{Pairing and uncertainty}

Runs are paired by task instance and repetition index.  To account for three
repetitions nested within each instance, completion-rate sensitivity intervals
resample task instances rather than individual runs.  For each C-versus-baseline
contrast, we sample ten task instances with replacement, retaining all three
repetitions and their paired outcomes.  We generate 1,000,000 task-cluster
bootstrap samples with NumPy 2.4.4, PCG64 seed 20260810, and report 2.5th and
97.5th percentiles of the completion-rate difference.

Only ten instance clusters and two generating templates are available.  These
intervals describe sensitivity within the frozen suite.  Ten clusters do not
support population-level intervals or a stand-alone significance claim for
real-world software-development workflows.  Run-level exact statistical tests are left
to the artifact appendix as exploratory diagnostics because they do not correct
for task-level dependence.

\section{Results}
\label{sec:results}
\subsection{Evidence integrity}

All 150 matrix cells have unique instance--condition--repetition identifiers and
unique evidence paths.  For each execution, the aggregate ledger records the
outcome, diagnostic score, token counts, execution-window and archived harness times, and
evidence location.  We rechecked the internal study archive: all 150 raw JSON
paths resolved, their byte counts and SHA-256 digests
matched the manifest, and recomputation of the included CSV fields found no
mismatch.  The public-artifact boundary is stated in \cref{sec:artifact-status}.

As noted in \cref{sec:method}, D and E were collected after A--C.  Comparisons
involving D or E may therefore include provider drift and are interpreted
descriptively throughout this section.

\subsection{RQ1: passing executions}

\Cref{tab:overall} gives the aggregate outcomes.  In the primary A--C batch,
condition C passes 28/30 executions (93.3\%), compared with 15/30 for A and
16/30 for B.  Later-batch D and E pass 15/30 and 14/30.  Every task instance passes at least once under C; only five
to seven instances do so under the alternatives.

\begin{table}[t]
  \centering
  \caption{Aggregate outcomes and accounted model tokens.  The final column
  pools every valid attempt and divides by the observed passes.  D/E were
  collected in a later non-interleaved batch.}
  \label{tab:overall}
  \small
  \begin{tabularx}{\linewidth}{@{}p{0.07\linewidth}Yrrrr@{}}
    \toprule
    ID & Condition & Passes /30 & Instances $\geq$1 pass /10 & Tokens/attempt & Attempt-accounted tokens/pass \\
    \midrule
    A & Re-investigation & 15/30 & 5/10 & 59,268 & 118,537 \\
    B & Full history & 16/30 & 7/10 & 67,726 & 126,986 \\
    C & \system{} handoff & \textbf{28/30} & \textbf{10/10} & 65,616 & \textbf{70,303} \\
    D & Rolling summary & 15/30 & 5/10 & 60,763 & 121,526 \\
    E & Lightweight JSON & 14/30 & 5/10 & 60,083 & 128,749 \\
    \bottomrule
  \end{tabularx}
\end{table}

The aggregate must be read together with the descriptive family split
(\cref{tab:family-results}).  On S, where all decision material arrives in one
stage, every condition passes 14 or 15 of 15 executions, placing C in the same
observed pass-rate range.  On GAR, where the release scope is carried through
three evidence updates, primary-batch C passes 14/15, B passes 2/15, and A
passes none; later-batch D and E also pass none.  The observed pass counts
therefore diverge on the GAR template.  Because S and GAR are not matched
instances, the split does not identify cross-stage dependence as the cause.

\begin{table}[t]
  \centering
  \caption{Results stratified by task family.  A dash indicates that no
  execution passed, so attempt-accounted cost per pass is undefined.}
  \label{tab:family-results}
  \small
  \begin{tabular}{@{}llrrr@{}}
    \toprule
    Family & ID & Passes /15 & Tokens/attempt & Attempt-accounted tokens/pass \\
    \midrule
    S & A & 15/15 & 33,792 & 33,792 \\
      & B & 14/15 & 33,809 & 36,224 \\
      & C & 14/15 & \textbf{20,784} & \textbf{22,268} \\
      & D & 15/15 & 33,742 & 33,742 \\
      & E & 14/15 & 33,723 & 36,132 \\
    \addlinespace
    GAR & A & 0/15 & 84,745 & -- \\
        & B & 2/15 & 101,643 & 762,319 \\
        & C & \textbf{14/15} & 110,449 & \textbf{118,338} \\
        & D & 0/15 & 87,784 & -- \\
        & E & 0/15 & 86,443 & -- \\
    \bottomrule
  \end{tabular}
\end{table}

A task-instance bootstrap yields positive descriptive intervals for the frozen
five-S/five-GAR mixture (\cref{tab:bootstrap}).  Because the suite contains only
two generating templates, the intervals do not quantify uncertainty over task
families and should not be interpreted through whether an endpoint crosses
zero.  For D and E, the resampling also does not model possible provider drift
from their later, non-interleaved collection batch.

\begin{table}[t]
  \centering
  \caption{C-minus-baseline completion-rate differences with paired
  task-cluster bootstrap sensitivity intervals.}
  \label{tab:bootstrap}
  \small
  \begin{tabular}{@{}lrrr@{}}
    \toprule
    Contrast & C & Baseline & Difference [95\% bootstrap sensitivity interval] \\
    \midrule
    C vs. A & 93.3\% & 50.0\% & 43.3 [10.0, 76.7] pp \\
    C vs. B & 93.3\% & 53.3\% & 40.0 [13.3, 66.7] pp \\
    C vs. D & 93.3\% & 50.0\% & 43.3 [10.0, 76.7] pp \\
    C vs. E & 93.3\% & 46.7\% & 46.7 [16.7, 76.7] pp \\
    \bottomrule
  \end{tabular}
\end{table}

\subsection{RQ2: accounted model-token efficiency}

Across all valid attempts, C uses 70,303 accounted model tokens per observed
pass.  Relative to A, B, D, and E, this attempt-accounted ratio is lower by
40.7\%, 44.6\%, 42.2\%, and 45.4\%, respectively.  It is a yield-adjusted
benchmark ratio: C averages 65,616 tokens per attempt, compared with 59,268 for
A, 67,726 for B, 60,763 for D, and 60,083 for E.  C's lower cost per observed
pass coincides with its substantially higher GAR yield; per-attempt token use has
no consistent advantage.

The family split clarifies how the aggregate ratio arises.  Causal attribution
requires additional controls.  On S, where no handoff is required, C uses 20,784 tokens per attempt
versus 33,723--33,809 for the alternatives while achieving comparable pass
rates.  C uses one Workbench agent turn for an S attempt, whereas each direct
condition uses a discovery turn followed by a decision turn.  The provider's
lower-level model-call count was not exposed.  The S token gap
therefore exposes a runner and prompt-structure difference beyond cross-stage
state availability and cannot be credited to the handoff mechanism.  On GAR, a
C attempt uses 110,449 tokens on average---more than A, D, or E---but
14 attempts pass.  The only GAR baseline with a passing result is B, whose two
passes imply 762,319 tokens per passing execution versus C's 118,338.  The GAR
comparison therefore shows a lower completion-adjusted token ratio for C; its per-attempt prompt
is not compressed.

In the descriptive both-passing subsets defined in \cref{sec:method}, the
median paired relative difference shows 38.75\%, 38.59\%, 38.66\%, and
38.78\% fewer tokens for C than for A, B, D, and E.  These pairs are dominated by S: all 14
matched passes against A, D, and E are S executions, while the 16 matches
against B contain 14 S and two GAR executions.  The 38.6\%--38.8\% range therefore
describes this predominantly single-stage subset and does not establish a general
multi-stage compression effect.

Diagnostic telemetry is directionally consistent with less work spent recovering
earlier information.  C records no rereads used to recover earlier information on average, compared with 0.93 for A, 0.53 for B,
0.47 for D, and 0.33 for E; mean total reads are 4.00 for C and 4.23--4.80 for
the alternatives.  The read-count difference is secondary evidence because the
interfaces expose different ways to obtain prior information.

\subsection{RQ3: condition-specific time trade-off}

\Cref{tab:timing-overall} reports both recorded time fields.  C averages
73.8 s execution-window time and 78.5 s in the archived harness-time field per attempt,
more than every
alternative.  Pooling all attempt time and dividing by observed passes gives C
79.1/84.1 s per pass.  The execution-window ratios are lower than A and B because their
failed GAR attempts are charged to few passes, not because a C attempt is
faster.

These measurements compare bundled conditions under their documented
condition-specific windows; they do not estimate isolated AstronOS runtime
overhead.

\begin{table}[t]
  \centering
  \caption{Overall time in seconds.  Per-pass columns charge every valid
  attempt to the observed passes.}
  \label{tab:timing-overall}
  \small
  \begin{tabular}{@{}lrrrr@{}}
    \toprule
  ID & Execution window/attempt & Archived harness field/attempt & Execution window/pass & Archived harness field/pass \\
    \midrule
    A & 42.3 & 42.3 & 84.7 & 84.7 \\
    B & 57.2 & 57.2 & 107.2 & 107.2 \\
    C & 73.8 & 78.5 & 79.1 & 84.1 \\
    D & 33.2 & 33.2 & 66.4 & 66.4 \\
    E & 32.1 & 32.1 & 68.9 & 68.9 \\
    \bottomrule
  \end{tabular}
\end{table}

Among matched executions in which both conditions pass, C's median execution-window
time is 2.0\%, 8.8\%, 34.9\%, and 48.9\% higher than A, B, D, and E.  Using the
condition-specific archived harness field, the corresponding increases are
27.4\%, 32.4\%, 68.7\%, and 86.3\%.  These subsets contain 14, 16, 14, and 14 pairs and are almost
entirely S executions, so the percentages apply to the both-passing subset,
which is predominantly S.  C's archived harness field exceeds its execution-window time
by 4.68 s on average.  Because the harness timers have condition-specific
boundaries, that difference is only a diagnostic and does not isolate platform
overhead.  The execution-window result is a descriptive trade-off whose
scope is the current both-passing subset.

\subsection{Failure analysis}

The family-level outcome is more informative than a single average.  In the
example cells \path{GAR-101-r1-A}, \path{GAR-101-r1-D}, and
\path{GAR-101-r1-E}, stage 1 is correct, but stages 2 and 3 contain no saved
structured decision, the final simulated submission is absent, and each run
performs two rereads to recover earlier information.  Each receives 15/100 diagnostic points.  The D/E
examples show that the frozen direct runner failed to turn the retained fields
into the required later-stage actions.  The effect of prose versus JSON remains
unidentified.  B
occasionally carries enough history to finish but does so in only 2/15
executions.

C has two failures.  GAR-103-r1-C reaches all three stages and preserves
the checkpoint correctly, but submits a non-canonical dependency order; its
S-102-r2-C likewise selects the correct release and hold sets but orders the
changes incorrectly.  Both receive high diagnostic scores (90 and 85) yet fail
the binary criterion.  Task-specific reasoning therefore remains necessary even
when state continuity is preserved.

The component scores are similarly concentrated.
All 15 C runs receive full credit for GAR classification, evidence, progression,
checkpoint handling, and output completeness; 14 receive full ordering credit.
Only the two passing B runs receive full credit in any of those categories, and
no A, D, or E run does.  The appendix reports these counts without inferring
which bundled treatment feature caused them.

The result pattern supports a narrow conclusion.  C provides no pass-rate
benefit on the single-stage reference family, where handoff is unnecessary.  The tested
AstronOS strategy has a large observed association with passing outcomes on this
three-stage template, alongside more per-attempt work and higher
both-passing time.  The contrast between the two observed family-level
patterns is the central empirical finding; because the families are not
matched, it is not an estimate of a causal interaction with task structure.

\section{Discussion}
\label{sec:discussion}
\subsection{Answers to the research questions}

\textbf{RQ1: scorer pass rate.}  The observed advantage is specific to the
cross-stage template.  In the primary batch, C passes 14/15 GAR executions,
B passes 2/15, and A passes none; later-batch D and E also pass none.  On the S
reference family, all conditions pass at least 14/15.  The
tested bundled \system{} strategy is associated with a higher pass rate on
this template.  Whether the handoff alone causes the effect and whether the
result generalizes across single-stage tasks remain open questions.

\textbf{RQ2: accounted token efficiency.}  C has the lowest total accounted
model tokens per observed pass across the whole matrix.  Two different
effects produce that aggregate: on S, C uses fewer tokens per attempt at a
similar pass rate; on GAR, it uses more tokens per attempt but turns nearly every
attempt into a passing result.  The second effect is yield-adjusted efficiency;
C's prompt text itself is not shorter.

\textbf{RQ3: time.}  Across all both-passing comparisons, C has the longer
median execution-window time.  Its attempt-accounted execution-window time per observed pass is
nevertheless below A and B because their non-passing attempts are charged to
relatively few passes.  The tested strategy trades per-attempt time for outcome
yield and attempt-accounted token cost.

\subsection{What the experiment says about the execution model}

The experiment compares complete prior-state handoff strategies rather than
isolating any single runtime component.  Its most relevant descriptive pattern is
little outcome difference on the single-stage reference family but a large
difference on the distinct three-stage template, where a later stage must carry
an exact scope through new evidence and a deterministic checkpoint.  This
pattern is consistent with the model's
premise that the enduring work item carries cross-step continuity while a
session serves as an execution context.

C uses the implemented Case, Task, and Scenario execution path from the pinned
AstronOS source revision and provides later stages with a runtime-admitted
previous-stage result, structured decision artifacts, and hash-pinned stage packages.  A stronger
causal account requires holding the full runner fixed while removing or altering
one element at a time.

The code and trace audit serve a different purpose.  They show that persistent
Case identity, snapshot revision, result fencing, Scenario interpretation, and
source-bearing product context are implemented mechanisms observed in the pinned revision.  Their evidentiary
role is to confirm correspondence between the paper's description and the implementation; component-level causal attribution
requires a different experiment.  The model layer defines an interface, the system
layer demonstrates concrete mechanisms and boundaries, and the empirical layer
measures one complete state-handoff strategy.

\subsection{When the runtime cost is likely to be worthwhile}

The S reference family provides a descriptive lower-bound case.  When all
relevant facts arrive together, the five observed pass counts differ by at most
one, while the lighter conditions have lower latency.  This pattern may favor a
summary or JSON record for tasks with the same structure, but establishing a
general default for one-shot work requires matched tasks beyond this family.

The S--GAR split is descriptive, not a matched staging intervention: the two
families use different generators, task sizes, and evidence structures.  The
GAR template contains multiple stages, state revision, an exact intermediate
scope, and a final decision based on the runtime-admitted scope after earlier
guesses have been superseded.  On this template, bundled condition C includes
an additional runtime transition and has a higher observed pass count.  Because
S and GAR are unmatched and C changes several factors together, the design does
not identify that transition or cross-step dependence as the cause.  It instead
motivates matched evaluations of state-dependent work in broader workflows.

This distinction also refines the meaning of efficiency.  A system can be slower
for a successful attempt yet cheaper per passing outcome if it greatly improves
yield.  Conversely, attempt-accounted cost per observed pass can look poor when the benchmark
contains many failures even if each failed attempt stops quickly.  Reporting
pass rate, per-attempt cost, attempt-accounted cost per pass, and
both-passing time together is therefore more informative than a single
speedup number.

\subsection{Alternative explanations}

Four alternatives remain plausible.  First, C may expose prior state in a form
that is simply easier for this model to follow.  That property is part of the
complete interface being evaluated.  Isolating versioning or Scenario progression
requires additional controls.  Second, the deterministic D and E projections
may omit packaging information present in C's runtime-admitted result and
decision artifact.  Publishing the exact content supplied in each condition is
necessary for readers to assess whether information coverage is fair.

Third, the output schema and scorer reward exact structured decisions.  All
conditions face the same checks, but a runtime that already represents state
structurally may align better with that construct.  Tasks with qualitative or
open-ended outcomes may show a different pattern.  Fourth, capacity-invalidated
attempts and non-model platform work are excluded from the primary cost totals.
The reported accounting covers model costs within this benchmark; a complete operational
cost model requires additional platform telemetry.

\subsection{Next experiments}

The highest-value extension is an equal-information control that supplies C's exact model-visible payload
through the direct runner, followed by factorial ablations within one runner:
product Context Bundle on/off, structured runtime-admitted result versus prose,
explicit state binding versus conversation-only continuation, and
Scenario-mediated versus direct progression.
A second priority is to vary the number of stages and the amount of superseded
history to locate the break-even point between runtime overhead and avoided
failure.  A third is external breadth: additional task families, models, and
    realistic but privacy-safe software-development workflows.  A complete
lightweight runtime baseline would further distinguish the value of the unified
model from the value of a particular implementation.

\section{Threats to Validity, Reproducibility, and Ethics}
\label{sec:validity}
\subsection{Internal and construct validity}

The five conditions share the generated tasks, stage-local evidence, requested
model identifier and reasoning setting, synthetic tools, and deterministic
scorer.  Their prior inputs differ in semantic coverage, representation, prompt,
and orchestration.  The frozen comparison is internally consistent at the
bundled-strategy level; separating information coverage, format, turn structure,
and runtime progression requires additional controls.  C
uses the implemented \system{} runtime path on synthetic task material, whereas the direct conditions use a common
lightweight runner.  The one-versus-two-Workbench-turn difference on S confirms that
execution structure differs even when no prior-state handoff is required.

Passing is an exact, executable construct for the synthetic staged decision-continuity benchmark.
The scorer is deterministic and frozen, which removes judge-model variance, and
emphasizes classification, dependency order, evidence, stage state, and final
simulated submission.  Maintainability, user satisfaction, and employee
productivity lie outside this construct.
The binary criterion intentionally marks high-scoring but non-canonical outputs
as failures.

Token accounting covers provider-reported model usage archived by the
experiment ledger.  It excludes database, CPU, network, and unobserved
control-plane work, as well as capacity-invalidated attempts.  Condition-specific execution-window
time and the condition-specific archived harness-time field are both reported,
but component-level overhead is not measured.
Conditions D and E were collected after A--C.  Because the provider snapshot,
seed, and sampling controls were unavailable, cross-batch drift can affect pass
rate, token use, and time even though the paper environment itself was isolated.  These
boundaries preclude claims about full platform cost or universal wall-clock
speed.

As a lower-bound sensitivity check, charging the observable archived token use
of the three capacity-invalidated attempts to their intended conditions yields
120,837 tokens/pass for A, at least 121,526 for D, and 131,224 for E; the D
attempt has incomplete token telemetry.  C remains 70,303 and is below all
three values or lower bounds.  Charging the archived execution-window time as
well yields 89.70 s/pass for A, 68.59 for D, and 77.85 for E, compared with
C's unchanged 79.06.  The compact manifest records the two D/E exclusions,
while the full three-attempt retry audit remains in the internal archive.

\subsection{External and statistical validity}

Ten parameterized instances derive from only two synthetic templates, and all
use one model configuration.  The three repetitions within an instance are
correlated stochastic replications.  Task-cluster bootstrap intervals preserve
that nesting, but ten clusters are too few to represent a population of
real-world software-development workflows; the shared S and GAR generators
further reduce effective diversity.
We therefore report effect sizes and descriptive intervals.  Ten task clusters
support descriptive sensitivity analysis, not a general significance claim.

The GAR template deliberately concentrates cross-stage dependencies and exact
state reuse.  This makes it diagnostic for the proposed boundary but may make it
more favorable to the tested \system{} strategy than loosely coupled work.  The
single-stage reference family provides a descriptive counterpoint but does not
test this concern under matched task content.  Generalization requires task families with different kinds of
state revision, tool use, and outcome semantics.

\subsection{Reproducibility and artifact status}
\label{sec:artifact-status}

The paper records the pinned source revision, requested model identifier and
reasoning setting, run matrix, per-run aggregate table, evidence manifest, and
verification script.  The internal result directory contains 150 result rows and
hashes that resolve to the archived raw evidence in the study environment.  We
rechecked that archive and found no missing raw file, size mismatch, digest
mismatch, or CSV-to-raw-field mismatch.

The compact public artifact supports arithmetic recomputation of the 150-row
ledger only.  It does not permit independent reconstruction of the tasks,
execution of the scorer, comparison of model-visible information across
conditions, or verification of the narrated raw traces.  It includes aggregate
JSON, an evidence manifest, and a verifier for matrix integrity and core
aggregate and pairwise ledger summaries.
Raw-evidence portability requires replacing manifest paths that currently
resolve only inside the internal study archive and sanitizing transcript paths
before release.  A portable public artifact must add
the synthetic task definitions, five condition builders, deterministic scorer,
anonymized raw results, retry ledger, and relative-path cryptographic manifest.
The headline outcome, token, time, and per-task tables can be recomputed from the
included CSV.  The diagnostic-category table and specific trace narratives rely
on the internal raw records.  Full rerun reproducibility also depends on a
public code artifact and continued access to the named external model.

\subsection{Ethics and data handling}

The evaluation uses no human subjects and no real customer, employee, source
repository, or business-system data.  All entities and tool effects are
synthetic.  The isolated paper environment did not write to the shared
integration environment.  Model-service credentials and internal network
details are excluded from manuscript and public artifacts.  The measured
construct is the runtime organization of work; workplace productivity and changes in
human--AI responsibility remain outside the evaluation.

\section{Conclusion}
\label{sec:conclusion}
We introduced a unified execution model for long-horizon agentic work.  The
model treats an enduring work item and its accepted state as the execution
boundary, constructs each step from an identified state basis, and separates an
executor's candidate output from the commit that advances authoritative state.
\system{} realizes these abstractions on audited paths through Cases, Tasks,
Scenario Packs, central orchestration, and local execution; source and trace audits show the
concrete mechanisms and their current boundaries.

In 150 valid executions over two templates of the synthetic staged decision-continuity benchmark, the
\system{} runtime-mediated state-handoff condition passes 14/15 three-stage
executions in the primary batch, compared with 0/15 for A and 2/15 for B;
later-batch D and E each pass 0/15, while all conditions pass
14/15 or 15/15 single-stage reference executions.  Across the full matrix, C has the
lowest attempt-accounted model tokens per observed pass, but it uses more time
within its condition-specific execution window on the predominantly single-stage
both-passing subset.  The tested
bundled \system{} strategy is therefore associated with a higher scorer pass
rate on this staged decision template at a measurable time cost.  The benchmark
uses pre-generated evidence and simulated tool effects; it does not evaluate
code modification, test execution, or real software release.  Accordingly,
the evidence applies to one bundled, benchmark-specific execution strategy.
Isolating the effects of durable state, versioning, and Context Bundles, and
evaluating \system{} as a complete platform, require further experiments.

\appendix
\section{Task and outcome detail}
\label{app:tasks}

\begin{table}[H]
  \centering
  \caption{Passing repetitions (out of three) for every task instance.}
  \label{tab:per-task}
  \begin{tabular}{@{}llrrrrr@{}}
    \toprule
    Instance & Family & A & B & C & D & E \\
    \midrule
    GAR-101 & GAR & 0 & 1 & 3 & 0 & 0 \\
    GAR-102 & GAR & 0 & 1 & 3 & 0 & 0 \\
    GAR-103 & GAR & 0 & 0 & 2 & 0 & 0 \\
    GAR-104 & GAR & 0 & 0 & 3 & 0 & 0 \\
    GAR-105 & GAR & 0 & 0 & 3 & 0 & 0 \\
    S-101 & S & 3 & 3 & 3 & 3 & 3 \\
    S-102 & S & 3 & 2 & 2 & 3 & 2 \\
    S-103 & S & 3 & 3 & 3 & 3 & 3 \\
    S-104 & S & 3 & 3 & 3 & 3 & 3 \\
    S-105 & S & 3 & 3 & 3 & 3 & 3 \\
    \bottomrule
  \end{tabular}
\end{table}

\section{Workbench-turn and prompt diagnostics}
\label{app:model-diagnostics}

\Cref{tab:call-diagnostics} reports family-level means over all 15 attempts in
each cell.  A turn is one agent turn recorded by the workbench ledger; the
provider's lower-level model-call count was not exposed.
Prompt characters count submitted prompt text before provider tokenization;
input and output columns are provider-reported tokens.  The single-stage S
reference family makes a runner difference visible: the direct conditions use two turns,
whereas C uses one.  On the three-stage GAR family, C uses three synthetic decision-stage Tasks and
one additional model-using checkpoint-propagation Task.  These differences are part of the
bundled strategies being compared, rather than evidence about state
representation alone.

\begin{table}[H]
  \centering
  \caption{Mean workbench-turn and prompt diagnostics per valid attempt.}
  \label{tab:call-diagnostics}
  \scriptsize
  \setlength{\tabcolsep}{3.5pt}
  \begin{tabular}{@{}llrrrr@{}}
    \toprule
    Family & ID & Turns & Prompt chars & Input tokens & Output tokens \\
    \midrule
    S & A & 2.00 & 6,881 & 33,335 & 456 \\
      & B & 2.00 & 6,881 & 33,350 & 459 \\
      & C & 1.00 & 19,985 & 20,054 & 730 \\
      & D & 2.00 & 6,881 & 33,330 & 412 \\
      & E & 2.00 & 6,881 & 33,326 & 397 \\
    \addlinespace
    GAR & A & 4.80 & 17,117 & 83,494 & 1,251 \\
        & B & 5.00 & 34,594 & 100,183 & 1,459 \\
        & C & 4.00 & 100,054 & 107,064 & 3,385 \\
        & D & 5.00 & 16,865 & 86,612 & 1,172 \\
        & E & 4.93 & 16,684 & 85,237 & 1,206 \\
    \bottomrule
  \end{tabular}
\end{table}

\section{GAR scorer diagnostics}
\label{app:gar-diagnostics}

The deterministic scorer separates six requirements: correct classification,
dependency-valid order, pre-generated test-status evidence, correct stage progression,
checkpoint/control handling, and complete structured output.  A run receives
full credit for a category only when all checks in that category pass.
\Cref{tab:gar-categories} counts full-credit runs out of 15.  These category
counts locate the observed failures; they do not isolate which runtime component
caused them.

\begin{table}[H]
  \centering
  \caption{GAR executions receiving full credit in each diagnostic category
  (out of 15).}
  \label{tab:gar-categories}
  \scriptsize
  \setlength{\tabcolsep}{3pt}
  \begin{tabular}{@{}lrrrrrr@{}}
    \toprule
    ID & Classify & Order & Test status & Stages & Checkpoint & Output \\
    \midrule
    A & 0 & 0 & 0 & 0 & 0 & 0 \\
    B & 2 & 2 & 2 & 2 & 2 & 2 \\
    C & 15 & 14 & 15 & 15 & 15 & 15 \\
    D & 0 & 0 & 0 & 0 & 0 & 0 \\
    E & 0 & 0 & 0 & 0 & 0 & 0 \\
    \bottomrule
  \end{tabular}
\end{table}

\section{Latency detail}
\label{app:latency}

\begin{table}[H]
  \centering
  \caption{Family-level time in seconds.  Correct-run time averages only
  passing runs; per-pass columns include all valid attempts in the numerator.}
  \small
  \begin{tabular}{@{}llrrr@{}}
    \toprule
    Family & ID & Correct-run execution window & Execution window/pass & Archived harness field/pass \\
    \midrule
    S & A & 24.24 & 24.24 & 24.24 \\
      & B & 28.20 & 30.85 & 30.85 \\
      & C & 23.83 & 25.62 & 30.34 \\
      & D & 18.34 & 18.34 & 18.34 \\
      & E & 18.59 & 19.61 & 19.61 \\
    \addlinespace
    GAR & A & -- & -- & -- \\
        & B & 69.97 & 641.47 & 641.47 \\
        & C & 124.73 & 132.49 & 137.80 \\
        & D & -- & -- & -- \\
        & E & -- & -- & -- \\
    \bottomrule
  \end{tabular}
\end{table}

\section{Implementation evidence anchors}
\label{app:code}

The conformance audit uses the following source anchors at revision
\path{0b0349d5290ce6d0949c6e57b646c053d437e2b6}:

\begin{itemize}
  \item Case and Task contracts: \path{packages/contracts/src/case.ts} and
  \path{packages/contracts/src/task.ts}.
  \item Snapshot revision and compare-and-set merge:
  \path{packages/db/src/repositories/case-snapshots-repository.ts}.
  \item Analysis revision binding and outcome fencing:
  \path{packages/workflow-shared/src/activities/analysis-activity.ts} and
  \path{packages/workflow-shared/src/activities/analysis-outcome-activity.ts}.
  \item Product Context Bundle lineage:
  \path{packages/contracts/src/context-memory.ts} and
  \path{packages/db/src/repositories/context-bundles-repository.ts}.
  \item Edge-result fencing and Scenario interpretation:
  \path{apps/hub-api/src/services/task-result-terminal.ts}.
  \item Experiment-C boundary:
  \path{scripts/paper-experiments/pilot/astronos-condition-controller.ts} and
  \path{scripts/paper-experiments/pilot/astronos-evidence-projector.ts}; these checks require product Context
  Bundle and memory injection to remain disabled.
\end{itemize}

The compact manifest records the pinned source revision and, for each run, its
task, condition, repetition index, original archive path, byte count, and
SHA-256 digest.  Source paths above are revision-relative; the compact manifest
does not contain source line numbers or test anchors.

\bibliographystyle{unsrtnat}
\bibliography{references}

\end{document}